\documentclass[10pt,twocolumn,letterpaper]{article}

\usepackage{iccv}
\usepackage{times}
\usepackage{epsfig}
\usepackage{graphicx}
\usepackage{amsmath}
\usepackage{amssymb}

\usepackage[ruled]{algorithm2e}
\usepackage{enumitem}

\DeclareMathOperator*{\argmin}{arg\,min}

\usepackage[breaklinks=true,bookmarks=false]{hyperref}

\iccvfinalcopy 

\def\iccvPaperID{1826} 

\ificcvfinal\fi

\begin{document}

\title{Feature Weighting and Boosting for Few-Shot Segmentation}

\author{Khoi Nguyen and Sinisa Todorovic\\
Oregon State University\\
Corvallis, OR 97330, USA\\
{\tt\small {\{nguyenkh,sinisa\}}@oregonstate.edu}
}

\maketitle
\ificcvfinal\thispagestyle{empty}\fi

\begin{abstract}
This paper is about few-shot segmentation of foreground objects in images. We train a CNN on small subsets of training images, each mimicking the few-shot setting. In each subset, one image serves as the query and the other(s) as support image(s) with ground-truth segmentation. The CNN first extracts feature maps from the query and support images. Then, a class feature vector is computed as an average of the support's feature maps over the known foreground. Finally, the target object is segmented in the query image by using a cosine similarity between the class feature vector and the query's feature map.  We make two contributions by: (1) Improving discriminativeness of features so their activations are high on the foreground and low elsewhere; and (2) Boosting inference with an ensemble of experts guided with the gradient of loss incurred when segmenting the support images in testing.   Our evaluations on the PASCAL-$5^i$ and COCO-$20^i$ datasets demonstrate that we significantly outperform existing approaches.
\end{abstract}

\begin{figure}[t!]
    \centering
    \includegraphics[scale=0.41]{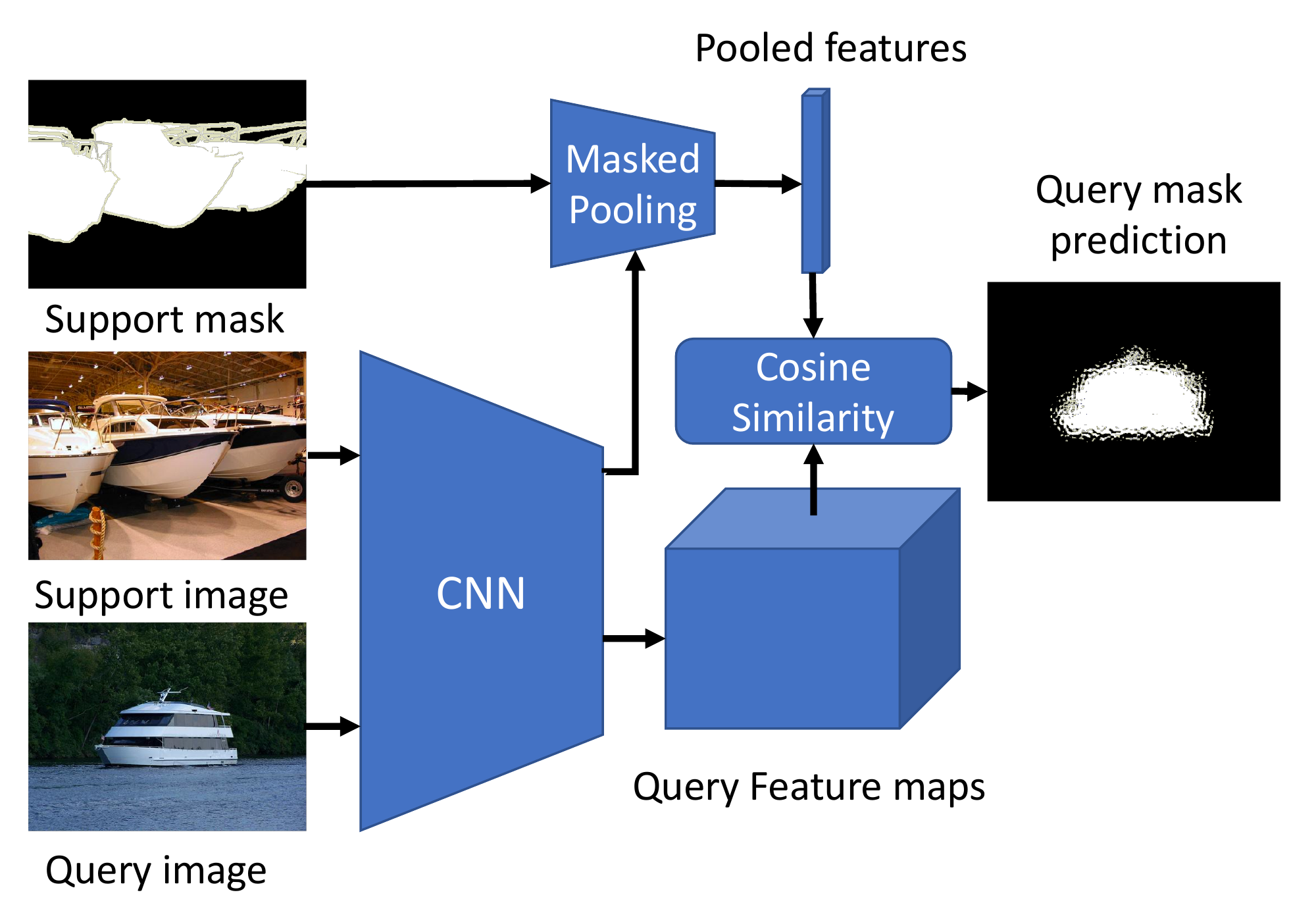}
    \caption{Our goal is to segment a foreground object class in a query image, given a single (few) support image(s) showing the same class with the known ground-truth segmentation mask(s). The target class may appear in the query and support images in  different numbers of instances and 3D poses. Recent work uses cosine similarity to first estimate a similarity map between CNN features of the query and support images. The similarity map is then used for segmentation of the target class in the query image.}
    \label{fig:intro}
\end{figure}

\section{Introduction}

 This paper is about few-shot segmentation of foreground objects in images. As  Fig.~\ref{fig:intro} shows, given only a few training examples -- called support images -- and their ground-truth segmentation of the target object class, our goal is to segment the target class in the query image. 
This problem is challenging, because the support and query images may significantly differ in the number of instances and 3D poses of the target class, as illustrated in Fig.~\ref{fig:intro}. This important problem arises in many applications dealing with scarce training examples of target classes.

 Recently, prior work has addressed this problem by training an object segmenter on a large training set, under the few-shot constraint \cite{shaban2017one,dong2018few,rakelly2018conditional}. The training set is  split into many small subsets. In every subset, one image serves as the query and the other(s) as the support image(s) with known ground truth(s). As shown in Fig.~\ref{fig:intro}, their framework uses a CNN -- e.g.,  VGG \cite{simonyan2014deep} or ResNet \cite{he2015deep} -- for extracting feature maps from the support and query images. The support's feature maps are first pooled over the known ground-truth foreground. Then, the support's masked-pooled features are used to estimate a cosine similarity map with the query's features. The resulting similarity map and the query's features are finally passed to a few convolutional layers in order to segment the target object class in the query. The incurred loss between the prediction and the query's ground-truth is used for the CNN's training.

\begin{figure*}[h!]
    \centering
    \includegraphics[scale=0.5]{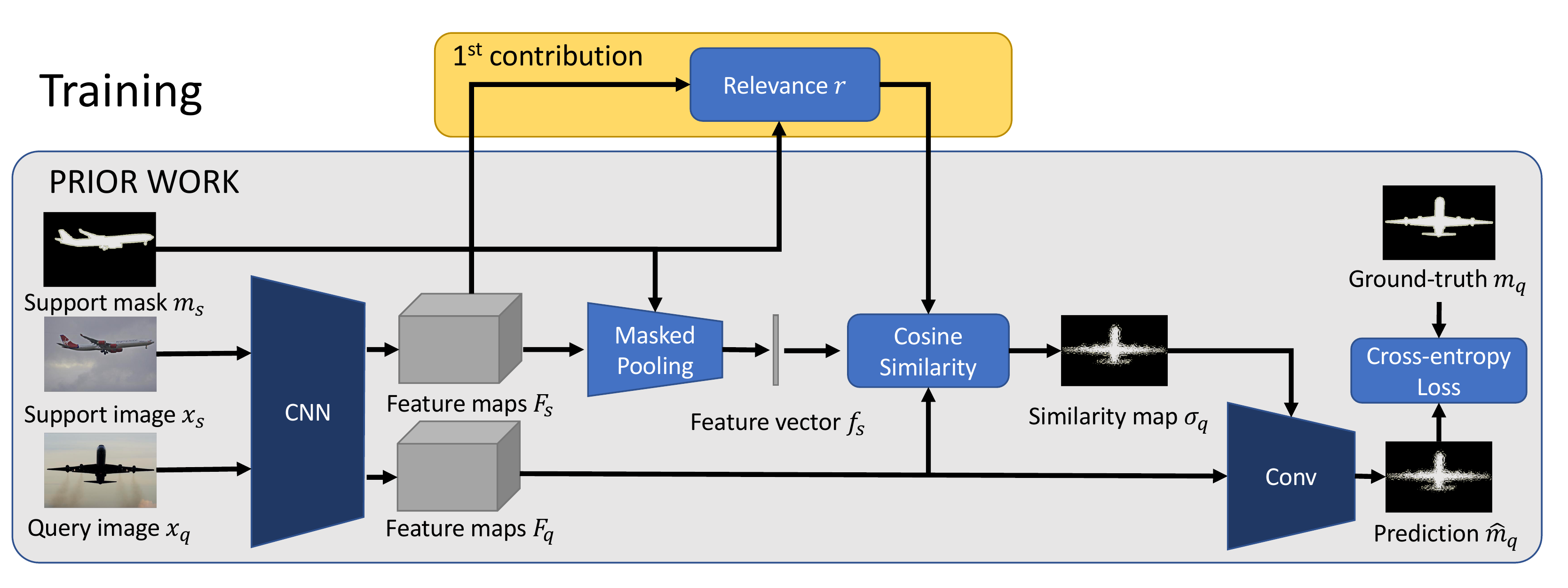}
    \caption{Our few-shot learning from a single support image: Prior work averages a class feature vector over the known foreground of the support image. A dot product of this class feature vector and feature maps of the query image gives a similarity map. The similarity map and features of the query image are used for segmenting the target class in the query. We extend prior work by additionally estimating feature relevance (contribution 1). Our contribution 2 is shown in Fig.~\ref{fig:test_phase}.}
    \label{fig:architecture}
\end{figure*}

The above framework has two critical limitations which we address in this paper. First, we experimentally found that the CNN has a tendency to learn non-discriminative features with high activations for different classes. To address this issue, as Fig.~\ref{fig:architecture} shows, our first contribution extends prior work by efficiently estimating feature relevance so as to encourage that their activations are high inside the ground-truth locations of the target class, and low elsewhere in the image. This is formulated as an optimization problem, for which we derive a closed-form solution.  
 
\begin{figure*}[h!]
    \centering
    \includegraphics[scale=0.52]{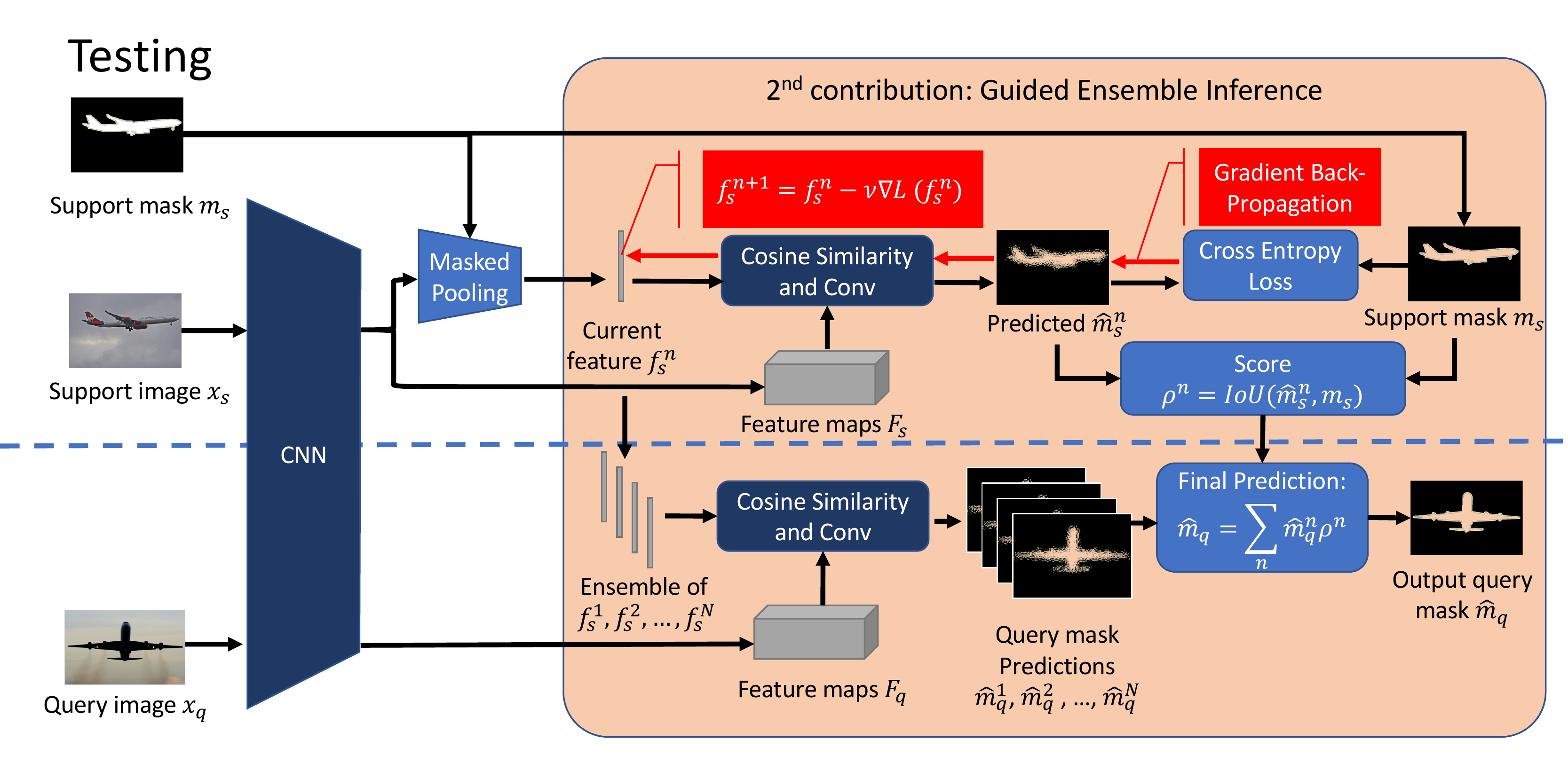}
    \caption{Our new boosted inference: We generate an ensemble of the support's features guided by the gradient of loss, incurred when segmenting the target object in the support image. A dot product between the query's feature maps and the ensemble of the support's features gives the corresponding ensemble of similarity maps, whose weighted average is taken as our final segmentation of the query.}
    \label{fig:test_phase}
\end{figure*}

Second, learning from few support images is prone to overfitting and poor generalization to the query in the face of the aforementioned large variations of the target class. To address this issue, as Fig.~\ref{fig:test_phase} shows, our second contribution is a new boosted inference, motivated by the traditional ensemble learning methods which are robust  to overfitting \cite{freund1999short,  friedman2001greedy}. We specify an ensemble of experts, where each expert adapts the features initially extracted from the support image. This feature adaptation is guided by the gradient of loss incurred when segmenting the support image relative to its provided ground truth. The ensemble of experts produce the corresponding ensemble of object segmentations of the query image, whose weighted average is taken as our final prediction. Importantly, while we use the first contribution in both training and testing, similar to the traditional ensemble learning methods, our second contribution is applied only in testing for boosting the performance of our  CNN-based segmenter. 

For $K$-shot setting, both contributions are naturally extended for segmenting the query image by jointly analyzing the provided support images and their ground-truths rather than treating support images independently as in prior work. 

For evaluation, we compare with prior work on the benchmark PASCAL-$5^i$ dataset \cite{shaban2017one}. Our results demonstrate that we significantly outperform the state of the art. In addition, we perform evaluation on the larger and more challenging COCO-$20^i$ dataset \cite{lin2014microsoft}. To the best of our knowledge, we are the first to report results of few-shot object segmentation on COCO-$20^i$.

In the following, Sec.~\ref{sec:related_work} reviews previous work, Sec.~\ref{sec:method} specifies our two contributions, Sec.~\ref{sec:implement} describes our implementation details and complexity, and Sec.~\ref{sec:experiments} presents our experimental results.

\section{Related Work}
\label{sec:related_work}
This section reviews related work on few-shot image classification and semantic segmentation. 

\textbf{Few-shot classification} predicts image class labels with access to few training examples. Prior work can be broadly divided into three groups: transfer learning of models trained on classes similar to the target classes  \cite{snell2017prototypical,ren2018meta,sung2018learning,vinyals2016matching}, meta-learning approaches that learn how to effectively learn new classes from small datasets \cite{finn2017model,nichol2018reptile}, and generative approaches aimed at data-augmentation  \cite{wang2018low,schwartz2018delta}. 

\textbf{Semantic segmentation} labels all pixels in the image. Recently, significant advances have been made by using fully convolutional network (FCN) \cite{shelhamer2016fully} and its variants --- including SegNet \cite{badrinarayanan2015segnet2}, UNet \cite{ronneberger2015unet}, RefineNet \cite{lin2016refinenet}, PSPNet \cite{zhao2016pyramid}, DeepLab v1\cite{chen2014semantic}, v2 \cite{chen2016deeplab}, v3 \cite{chen2017rethinking}, v3+ \cite{chen2018encoderdecoder} --- all of which are usually evaluated on the PASCAL VOC 2012 \cite{Everingham10} and MSCOCO \cite{lin2014microsoft} datasets. However, these approaches typically  require very large training sets, which limits their application to a wide range of domains. 

\textbf{Few-shot semantic segmentation} labels pixels of the query image that belong to a target object class, conditioned by the ground-truth segmentation masks of a few support images. Prior work typically draws from the above mentioned approaches to few-shot image classification and semantic segmentation. For example, the one-shot learning method OSLSM \cite{shaban2017one} and its extensions --- namely, Co-FCN \cite{rakelly2018conditional,rakelly2018fewshot},  PL+SEG \cite{dong2018few}, and SG-One \cite{zhang2018sgone} --- consist of the conditioning and segmentation branches implemented as  VGG \cite{simonyan2014deep} and FCN-32s \cite{shelhamer2016fully}, respectively. The conditioning branch analyzes the target class in the support image, and conditions the segmentation branch for object segmentation in the query image. Co-FCN improves OSLSM by segmenting the query image based on a concatenation of pooled features from the support image and feature maps from the query image.  PL+SEG first estimates a distance between the query's feature maps and prototypes predicted from the support image, and then labels pixels in the query image with the same class as their nearest neighbor prototypes. SG-One also estimates similarity between a pooled feature from the support image and feature maps  of the query for predicting the query's segmentation. Our approach extends SG-One with the two contributions specified in the next section.

\section{Our Approach}
\label{sec:method}

An object class is represented by $K$ support images with  ground-truth segmentation masks. Given a query image showing the same object class in the foreground, our goal is predict the foreground segmentation mask. For $K=1$, this problem is called one-shot semantic segmentation. Below, and in the following Sec.~\ref{sec:training} and Sec.~\ref{sec:ensemble}, we consider the one-shot setting, for simplicity. Then, we discuss the K-shot setting, $K > 1$, in Sec.~\ref{sec:k-shot}

Given a large set of training images showing various object classes and the associated ground-truth segmentation masks, our approach follows the common episodic training strategy. In each training episode, we randomly sample a pair of support and query images, $x_s$ and $x_q$, with binary segmentation masks, $m_s$ and $m_q$, of the target object class in the foreground. Elements of the mask are set to 1,  $m_{s,i}=1$, for pixels $i$ occupied by the target class; otherwise, $m_{s,i}=0$. The same holds for $m_q$. We use $m_s$ to condition the target class  in the query image. 

The standard cross-entropy loss, $L(\hat{m}_q, m_q)$, between the binary ground-truth $m_q$ and predicted query mask $\hat{m}_q=f_{\theta}(x_s, m_s, x_q) $, is used for the end-to-end training of parameters $\theta$ of our deep architecture, $\theta^* = \argmin_\theta L(\hat{m}_q, m_q)$.

\subsection{Training of Our Deep Architecture}\label{sec:training}
Fig.~\ref{fig:architecture} shows the episodic training of a part of our deep architecture (without our second contribution which is not trained) on a pair of support $x_s$ and query $x_q$ images. We first use a CNN to extract feature maps $F_s\in\mathbb{R}^{d\times w\times h}$ from $x_s$, and feature maps $F_q\in\mathbb{R}^{d\times w\times h}$ from $x_q$, where $d$ is the feature dimensionality, and $w$ and $h$ denote the width and height of the feature map. Then, we average $F_s$ over the known foreground locations in $m_s$, resulting in the average class feature vector $f_s$ of the support image. For this masked feature averaging, $m_s$ is down-sampled to $\tilde{m}_s\in\{0,1\}^{w\times h}$ with the size $w\times h$ of the feature maps, and $f_s$ is estimated as
\begin{equation}
  f_s = \frac{1}{|\tilde{m}_{s}|} \sum_{i =1}^{wh} F_{s,i}\; \tilde{m}_{s,i}.
  \label{eq:masked_pooling}
 \end{equation}
where $|\tilde{m}_{s}|=\sum_i \tilde{m}_{s,i}$ is the number of  foreground locations in $\tilde{m}_{s}$.
Next, we compute the cosine similarity between  $f_s$ and every feature vector $F_{q,i}$ from the query feature maps $F_{q}$. This gives a similarity map,  $\sigma_q\in[0,1]^{w\times h}$, between the support and query image:
\begin{equation}
\sigma_{q,i} = \cos(f_{s}, F_{q,i}) = \frac{f_s ^ T F_{q,i}}{\|f_s\|_2 \cdot \|F_{q,i}\|_2},\; i=1,\dots, w\; h.
\label{eq:similarity}
\end{equation}
We expect that $\sigma_q$  provides informative cues for object segmentation, as high values of  $\sigma_{q,i}$ indicate likely locations of the target class in the query image. 

Before we explain how to finally predict $\hat{m}_q$ from $\sigma_q$ and $F_q$, in the following, we specify our first technical contribution aimed at extending the above framework.

\vspace{5pt}

\noindent
{\bf Contribution 1: Feature Weighting.}
For learning more discriminative features of the target class from a single (or very few) support image(s), we introduce a regularization that encourages high feature activations  on the foreground and simultaneoulsy low  feature activations  on the background of the support image. 
This is formalized as an optimization problem for maximizing a sum of {\em relevant}   
differences between feature activations. Let $\phi_s{\in}\mathbb{R}^{d\times1}$ denote a vector of feature differences normalized over the foreground and background areas in the segmentation map of the support image as
\begin{equation}
\phi_s = \sum_{i=1}^{w h} F_{s, i}\left[\frac{\tilde{m}_{s,i}}{|\tilde{m}_{s}|} -  \frac{1-\tilde{m}_{s,i}}{w h-|\tilde{m}_{s}|}\right], 
\end{equation}
The relevance $r\in\mathbb{R}^{d\times1}$ of features in $\phi_s$ is estimated by maximizing a sum of the feature differences:%
\begin{equation}
\underset{r}{\text{maximize}} \quad \phi_s^\top\; r, 
\quad \text{s.t.}\quad \| r \|_2 = 1\;.
\label{eq:optimization1}
\end{equation}
The problem in \eqref{eq:optimization1} has a closed-form solution:
\begin{equation}
r^* = \frac{\phi_s}{\|\phi_s\|_2}.
\label{eq:solu1}
\end{equation}

We use the estimated feature relevance $r^*$ when computing the similarity map between the support and query images. Specifically, we modify the cosine similarity between $f_s$ and $F_q$, given by \eqref{eq:similarity}, as%
\begin{equation}
\sigma_{q,i}^* = \cos(f_{s}, F_{q,i}, r^*) = \frac{(f_s \odot r^*) ^ T (F_{q,i} \odot r^*)}{\|f_s \odot r^*\|_2\cdot \|F_{q,i} \odot r^*\|_2},
\label{eq:modified_similarity}
\end{equation}
where $\odot$ is the element-wise product between two vectors. 

Note that we account for feature relevance in both training and testing. As $r^*$ has a closed-form solution, it can be computed very efficiently. Also, the modification of the similarity map in \eqref{eq:modified_similarity} is quite simple and cheap to implement in modern deep learning frameworks. 

As shown in Fig.~\ref{fig:architecture}, in the final step of our processing, we concatenate $\sigma_q^*$ and $F_q$ together, and pass them to a network with only two convolutional layers for predicting $\hat{m}_q$.

\subsection{Contribution 2: Feature Boosting}
\label{sec:ensemble}

\begin{algorithm}[b]

\label{algo:1}
 \KwIn{$F_s$, $f_s$, $F_q$, $m_s$,  $\nu$}
 \KwOut{$\hat{m}_q$}
  \tcp{Guided ensemble of  experts}
 1. Initialize: $f_s^1 = f_s, E = \{\}, R = \{\}$\;
 2. \For {$n=1,\dots,N$}{
  		a.   $\sigma_{s,i}^*=\cos(f_s^n,F_{s,i}^n,r^*)$, $i=1,... wh$, as in \eqref{eq:modified_similarity} \;
  		b.
  		$\hat{m}_s^n = \text{Conv} (\sigma_{s}^*, F_s^n)$\;        
        c. $\rho^n = \text{IoU}(\hat{m}_s^n, m_s)$\;
        d. Compute cross-entropy loss: $L(\hat{m}_s^n, m_s)$\;
        e. $f_s^{n+1} = f_s^{n} - \nu\;\partial L(\hat{m}_s^n,m_s)/\partial f_s^n$\;
        f. $E = E \cup \{f_s^n\}, R = R \cup \{\rho^n\}$, 
  }
 \tcp{Inference using $E$ and $R$}
 4. \For {$n=1,\dots,N$}{
 a.  $\sigma_{q,i}^*=\cos(f_s^n,F_{q,i},r^*)$, $i=1,... wh$, as in \eqref{eq:modified_similarity} \;
  		b.
  		$\hat{m}_q^n = \text{Conv} (\sigma_{q}^*, F_q)$\;
 
  }
  5. $\hat{m}_q=\sum_{n=1}^N\; \hat{m}_q^n\;\rho^n$
 \caption{Guided Ensemble Inference in Testing}

\end{algorithm}

In testing, 
the CNN is supposed to address a new object class which has not been seen in training.
To improve generalization to the new class, in testing, we use a boosted inference -- our second contribution -- inspired by the gradient boosting \cite{friedman2001greedy}.  Alg.\ref{algo:1} summarizes our boosted inference in testing. Note that in testing parameters of the CNN and convolutional layers remain fixed to the trained values.

As shown in Fig.~\ref{fig:test_phase}, given a support image with ground-truth $m_s$ and a query image, in testing, we predict not only the query mask $\hat{m}_q$, but also the support mask $\hat{m}_s$ using the same deep architecture as specified in Sec.~\ref{sec:training}.  $\hat{m}_s$ is estimated in two steps. First, we compute a similarity map, $\sigma_s^*$, as a dot product between $f_s\odot  r^*$ and  $F_{s,i}\odot r^*$, as in \eqref{eq:modified_similarity}. Second, we pass $\sigma_s^*$ and $F_s$ to the two-layer convolutional network for predicting $\hat{m}_s$. Third, we estimate the standard cross-entropy loss 
$L(\hat{m}_s,m_s)$, and iteratively update the average class features as
\begin{equation}
    f_s^{n+1} = f_s^n -\nu \;\partial L(\hat{m}_s^n,m_s)/\partial f_s^n,
    \label{eq:update}
\end{equation}
where $\nu$ is the learning rate. $f_s^n$, $n=1,\dots,N$, are experts that we use for predicting the corresponding query masks  $\hat{m}_q^n$, 
by first estimating the similarity map $\sigma_{q,i}^*=\cos(f_s^n,F_{q,i},r^*)$, $i=1,... wh$, as in \eqref{eq:modified_similarity}, and then passing $\sigma_q^*$ and $F_q$ to the two-layer network for computing $\hat{m}_q^n$. 

Finally, we fuse the ensemble  $\{\hat{m}_q^n:n=1,\dots,N\}$ into the final segmentation, $\hat{m}_q$, as
\begin{equation}
    \hat{m}_q=\sum_{n=1}^N\; \hat{m}_q^n\;\rho^n,
\end{equation}
where $\rho^n$ denotes our estimate of the expert's confidence in correctly segmenting the target class, computed  as the intersection-over-union score between $\hat{m}_s^n$ and $m_s$:
\begin{equation}
    \rho^n=\text{IoU}(\hat{m}_s^n,m_s).
\end{equation}

\subsection{$K$-shot Setting}\label{sec:k-shot}
When the number of support images $K > 1$, prior work \cite{rakelly2018conditional,rakelly2018fewshot,dong2018few,zhang2018sgone} predicts $\hat{m}_q^k$ for each support image independently,  and then estimates $\hat{m}_q$ as an average over these predictions, $\hat{m}_q=\frac{1}{K}\sum_{k=1}^K\hat{m}_q^k$. In contrast, our two contributions can be conveniently extended to the K-shot setting so as to further improve our robustness, beyond the standard averaging over $K$ independent segmentations of the query. 

Our contribution 1 is extended by estimating relevance $r\in\mathbb{R}^{d\times1}$ of a more general difference vector of feature activations defined as
\begin{equation}
\phi_s =  \sum_{k=1}^{K} \sum_{i}^{w h} F_{s, i}^k\;\left[\frac{\tilde{m}_{s,i}^k}{|\tilde{m}_{s}^k|} -  \frac{1 - \tilde{m}_{s,i}^k}{w h - |\tilde{m}_{s}^k|}\right]
\end{equation}
Similar to \eqref{eq:optimization1} and \eqref{eq:solu1}, the optimal feature relevance has a closed-form sulution $r^* = \frac{\phi_s}{\|\phi_s\|_2}$. Note that we  estimate $r^*$ jointly over all $K$ support images, rather than as an average of independently estimated feature relevances for each support image. We expect the former (i.e., our approach) to be more robust than the latter.

Our contribution 2 is extended by having a more robust update of $f_s^{n+1}$ than in \eqref{eq:update}:
\begin{equation}
    f_s^{n+1} = f_s^n -\nu \;\partial \sum_{k=1}^K L(\hat{m}_s^k,m_s^k)/\partial f_s^n,
    \label{eq:updateK}
\end{equation}
where $L(\hat{m}_s^k,m_s^k)$ is the cross entropy loss incurred for predicting the segmentation mask $\hat{m}_s^k$ using the  unique vector $f_s^n$ given by \eqref{eq:updateK} for every support image $k$, as explained in Sec.~\ref{sec:ensemble}. Importantly, we do not generate  $K$ independent ensembles of experts $\{f_s^n:n=1,\dots,N\}_{k=1,K}$ for each of the $K$ support images. Rather, we estimate a single ensemble of experts more robustly over all $K$ support images, starting with the initial expert $f^1_s = \frac{1}{K} \sum_{k=1}^{K}f_s^k$. 

\section{Implementation Details and Complexity}
\label{sec:implement}
{\bf Implementation.} The CNN we use  is a modified version of either VGG-16 \cite{simonyan2014deep} or ResNet-101 \cite{he2015deep}. 
Our CNN has  the last two convolutional layers modified so as to have the stride equal to 1 instead of 2 in the original networks. This is combined with a dilated convolution to enlarge the receptive field with rates 2 and 4, respectively. So the final feature maps of our network have $\text{stride} = 8$, which is 1/8 of the input image size. For the two-layer convolutional network (Conv)  aimed at producing the final segmentation, we use a $3\times 3$ convolution with ReLU  and 128 channels, and a $1\times1$ convolution with 2 output channels -- background and foreground. It is worth noting that we do not use a CRF  as a common post-processing step \cite{krhenbhl2012efficient}. 

For implementation, we use Pytorch   \cite{paszke2017automatic}. Following the baselines \cite{zhang2018sgone,rakelly2018conditional,rakelly2018fewshot}, we pretrain the CNN on ImageNet \cite{ILSVRC15}. Training images are resized to $512\times 512$, while keeping the original aspect ratio. All test images keep their original size. Training is done with the SGD, learning rate $7e^{-3}$, batch size  8,  and in 10,000 iterations.    For the contribution 2, the number of experts $N$ is analyzed in Sec.~\ref{sec:experiments}. For updating $f_s^n$ in \eqref{eq:update},  we use Adam optimizer \cite{kingma2014adam} with   $\nu=1e^{-2}$.

{\bf  Complexity.} 
In training, prior work \cite{zhang2018sgone,rakelly2018conditional,rakelly2018fewshot}: (1) Uses a CNN with complexity $O(\text{CNN})$ for extracting features from the support and query images, (2) Computes the similarity map with  complexity $O(d\;w\;h)$, and (3) \cite{zhang2018sgone} additionally uses a convolutional network for segmenting the query with  complexity $O(\text{Conv})$. Note that  $O(\text{Conv})=O(d\;w\;h)$ as both the similarity map and convolutions in the Conv network are computed over feature maps with the size $d\times w\times h$. Also, note that $O(d\;w\;h)$ is significantly smaller than $O(\text{CNN})$. Our contribution 1 additionally computes the feature relevance using the closed-form solution with a linear complexity in the size of the feature maps  $O(d\;w\;h)$. Therefore, our total training complexity is equal to that of prior work \cite{zhang2018sgone,rakelly2018conditional,rakelly2018fewshot}: $O(\text{Train}) = O(\text{CNN})  +O(d\;w\;h)$.

In testing, complexity of prior work \cite{zhang2018sgone,rakelly2018conditional,rakelly2018fewshot} is the same as $O(\text{Train})$. Our contribution 2 increases complexity in testing by additionally estimating the ensemble of $N$ segmentations of the query image. Therefore, in testing, our complexity is  $O(\text{Test}) = O(\text{CNN})  + O(N\;d\;w\;h)$. Thus, in testing, we increase only the smaller term of the total complexity. For small $N$, we have that the first term $O(\text{CNN})$ still dominates the total complexity. As we show in  
Sec.~\ref{sec:experiments}, for $N=10$, we significantly outperform the state of the art, which justifies our slight increase in testing complexity.

\section{Experiments} \label{sec:experiments}
{\bf Datasets.} For evaluation, we use two datasets: (a) PASCAL-$5^i$ which combines images from the PASCAL VOC 2012 \cite{Everingham10} and Extended SDS  \cite{hariharan2011semantic} datasets; and (b) COCO-$20^i$ which is based on the MSCOCO dataset \cite{lin2014microsoft}. For PASCAL-$5^i$, we use the same 4-fold cross-validation setup as prior work  \cite{shaban2017one,rakelly2018conditional,dong2018few}. Specifically, from the 20 object classes in PASCAL VOC 2012, for each fold $i=0,...,3$, we sample five as test classes, and use the remaining 15 classes for training. Tab.~\ref{tab:voc} specifies our test classes for each fold of PASCAL-$5^i$. As in \cite{shaban2017one}, in each fold $i$, we use $1000$ support-query pairs of test images sampled from the selected five test classes.

\begin{table}
  \centering
  \begin{footnotesize}
  \begin{tabular}{l|l}
    \hline
    \hline
    \textbf{Dataset} & \textbf{Test classes} \\
    \hline
    PASCAL-5\textsuperscript{0} & aeroplane, bicycle, bird, boat, bottle \\
    PASCAL-5\textsuperscript{1} & bus, car, cat, chair, cow \\
    PASCAL-5\textsuperscript{2} & diningtable, dog, horse, motorbike, person \\
    PASCAL-5\textsuperscript{3} & potted plant, sheep, sofa, train, tv/monitor \\
    \hline
    \hline
  \end{tabular}
  \caption{Evaluation on PASCAL-$5^i$ uses the 4-fold cross-validation. The table specifies 5
  test classes used in each fold $i=0,...,3$. The remaining 15 classes are used for training.}\label{tab:voc}
  \vspace{-13pt}
  \end{footnotesize}
\end{table}

\begin{table}[b]
\setlength{\tabcolsep}{3pt}
\begin{footnotesize}
\begin{tabular}{rl|rl|rl|rl}
\hline 
\hline
\multicolumn{2}{c|}{COCO-$20^0$} & \multicolumn{2}{c|}{COCO-$20^1$} & \multicolumn{2}{c|}{COCO-$20^2$} & \multicolumn{2}{c}{COCO-$20^3$} \\ \hline
1       & Person            & 2      & Bicycle           & 3      & Car               & 4      & Motorcycle        \\
5       & Airplane          & 6      & Bus               & 7      & Train             & 8      & Truck             \\
9       & Boat              & 10     & T.light     & 11     & Fire H.      & 12     & Stop          \\
13      & Park meter     & 14     & Bench             & 15     & Bird              & 16     & Cat               \\
17      & Dog               & 18     & Horse             & 19     & Sheep             & 20     & Cow               \\
21      & Elephant          & 22     & Bear              & 23     & Zebra             & 24     & Giraffe           \\
25      & Backpack          & 26     & Umbrella          & 27     & Handbag           & 28     & Tie               \\
29      & Suitcase          & 30     & Frisbee           & 31     & Skis              & 32     & Snowboard         \\
33      & Sports ball       & 34     & Kite              & 35     & B. bat      & 36     & B. glove    \\
37      & Skateboard        & 38     & Surfboard         & 39     & T. racket     & 40     & Bottle            \\
41      & W. glass        & 42     & Cup               & 43     & Fork              & 44     & Knife             \\
45      & Spoon             & 46     & Bowl              & 47     & Banana            & 48     & Apple             \\
49      & Sandwich          & 50     & Orange            & 51     & Broccoli          & 52     & Carrot            \\
53      & Hot dog           & 54     & Pizza             & 55     & Donut             & 56     & Cake              \\
57      & Chair             & 58     & Couch             & 59     & P. plant      & 60     & Bed               \\
61      & D. table      & 62     & Toilet            & 63     & TV                & 64     & Laptop            \\
65      & Mouse             & 66     & Remote            & 67     & Keyboard          & 68     & Cellphone        \\
69      & Microwave         & 70     & Oven              & 71     & Toaster           & 72     & Sink              \\
73      & Fridge      & 74     & Book              & 75     & Clock             & 76     & Vase              \\
77      & Scissors          & 78     & Teddy        & 79     & Hairdrier        & 80     & Toothbrush     \\
\hline
\hline
\end{tabular}
\end{footnotesize}
\caption{Evaluation on COCO-$20^i$ uses the 4-fold cross-validation. The table specifies 20
  test classes used in each fold $i=0,...,3$. The remaining 60 classes are used for training.}
\label{tab:coco}
\end{table}

We create COCO-$20^i$ for evaluation on a more challenging dataset than PASCAL-$5^i$, since  MSCOCO has 80 object classes and its ground-truth segmentation masks have lower quality than those in PASCAL VOC 2012. To the best of our knowledge, no related work has reported one-shot object segmentation on MSCOCO. For evaluation on COCO-$20^i$, we use 4-fold cross-validation. From the 80 object classes in MSCOCO, for each fold $i=0,...,3$, we sample 20 as test classes, and use the remaining 60 classes for training.  Tab.~\ref{tab:coco} specifies our test classes for each fold of COCO-$20^i$. In each fold, we sample 1000 support-query pairs of test images from the selected 20 test classes. 

\begin{table}[t]
  \centering
  \begin{tabular}{llcccc}
    \hline
    \hline
   & \textbf{Phase} & \textbf{B} & \textbf{B+C1} & \textbf{B+C2} & \textbf{B+C1+C2}\\
    \hline
    VGG & Train & 171 & 172 & 171 & 172 \\
    & Test & 148  & 150 & 211 & 211 \\
    \hline
    ResNet & Train & 386 & 388 & 386 & 388 \\
    & Test & 268  & 280 & 360  & 360 \\
    \hline
    \hline
  \end{tabular}
  \caption{Training and test times per example in milliseconds, with 1 Nvidia 1080 Ti GPU on PASCAL-5\textsuperscript{i}, for different ablations indicated in the top row. Using C2 increases only the test time. Specifically, we use $K=10$ in our experiments}
  \label{tab:time}
  \vspace{-10pt}
\end{table}

\begin{figure}[h!]
    \centering
    \includegraphics[scale=0.42]{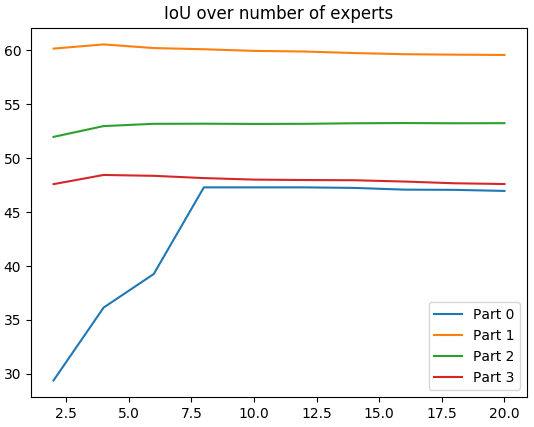}
    \caption{The mIoU of B+C1+C2 as a function of the number of experts $N$ in the one-shot setting on PASCAL-$5^i$. Part 0 -- Part 3 denote the four folds of cross-validation.}
    \label{fig:num_experts}
\end{figure}

\begin{table*}[t]
  \centering
  \begin{tabular}{l|l|cccc|c}
    \hline
    \hline
    \textbf{Backbone} & \textbf{Methods} & \textbf{PASCAL-5\textsuperscript{0}} & \textbf{PASCAL-5\textsuperscript{1}} & \textbf{PASCAL-5\textsuperscript{2}} & \textbf{PASCAL-5\textsuperscript{3}} & \textbf{Mean} \\
     \hline
	VGG 16 & OSLSM~\cite{shaban2017one} &  33.60 & 55.30 & 40.90 & 33.50 & 40.80\\
    & co-FCN~\cite{rakelly2018conditional} &  36.70  & 50.60 & 44.90 & 32.40 & 41.10 \\
    & PL+SEG+PT~\cite{dong2018few} &  -  & - & - & - & 42.70 \\
    & SG-One ~\cite{zhang2018sgone} & 40.20  & 58.40 & 48.40 & 38.40 & 46.30 \\
    \hline
    VGG 16& B &  42.39 & 57.43 & 47.84  & 42.55 & 47.55 \\
    & B + C1 & 43.34  & 56.72 & 50.64  & 44.01 & 48.68\\
    & B + C2 & 46.49 & \textbf{60.27} & 51.45  & 46.67 & 51.22 \\
    & B + C1 + C2 & \textbf{47.04 }& 59.64 & \textbf{52.61}  & \textbf{48.27} & \textbf{51.90} \\
    \cline{2-7}
    & Upper-bound & \textit{51.33} & \textit{64.20} & \textit{61.72}  & \textit{58.02} & \textit{58.82} \\
    \hline
    ResNet 101& B & 43.24  & 60.82 & 52.58  & 48.57  &  51.30 \\
    & B + C1 & 46.06  & 61.22 & 54.90  & 48.65 & 52.71 \\
    & B + C2 & 47.46  & 63.76 & 54.11  & 51.50 & 54.21 \\
    & B + C1 + C2 & \textbf{51.30} & \textbf{64.49} & \textbf{56.71}  & \textbf{52.24} & \textbf{56.19} \\
    \cline{2-7}
    & Upper-bound & \textit{61.34} & \textit{72.85} & \textit{70.20}  & \textit{67.90} & \textit{68.07} \\
    \hline
    \hline
  \end{tabular}
  \caption{Mean IoU  of one-shot segmentation on PASCAL-5\textsuperscript{i}. The best  results are in bold.}\label{tab:ablation}
\end{table*}

\begin{table*}[t]
  \centering
  \begin{tabular}{l|l|cccc|c}
    \hline
    \hline
     \textbf{Backbone} & \textbf{Methods} & \textbf{PASCAL-5\textsuperscript{0}} & \textbf{PASCAL-5\textsuperscript{1}} & \textbf{PASCAL-5\textsuperscript{2}} & \textbf{PASCAL-5\textsuperscript{3}} & \textbf{Mean} \\
    \hline
	VGG 16 & OSLSM~\cite{shaban2017one} &  35.90 & 58.10 & 42.70 & 39.10 & 43.95 \\
    & co-FCN~\cite{rakelly2018conditional} &  37.50  & 50.00 & 44.10 & 33.90 & 41.38 \\
    & PL+SEG+PT~\cite{dong2018few} &  -  & - & - & - & 43.70 \\
    & SG-One ~\cite{zhang2018sgone} & 41.90  & 58.60 & 48.60 & 39.40 & 47.10 \\
     \hline
    VGG 16 & Average  & 48.01  & 59.43 & 54.53  & 48.50 & 52.62 \\
    ResNet 101 & Average  & 51.72  & 64.99 & 61.05  & 53.34 & 57.78 \\
    \hline
    VGG 16 & Our-K-shot  & \textbf{50.87}  & \textbf{62.86} & \textbf{56.48}  & \textbf{50.09} & \textbf{55.08} \\
    ResNet 101 & Our-K-shot  & \textbf{54.84}  & \textbf{67.38} & \textbf{62.16}  &\textbf{55.30}& \textbf{59.92} \\
    \hline
    \hline
  \end{tabular}
  \caption{Mean IoU  of five-shot segmentation on PASCAL-5\textsuperscript{i}. The best results are in bold.}\label{tab:five-shot}
\end{table*}

\begin{table*}[t]
  \centering
  \begin{tabular}{l|l|cccc|c}
    \hline
    \hline
    \textbf{Backbone} & \textbf{Methods} & \textbf{COCO-20\textsuperscript{0}} & \textbf{COCO-20\textsuperscript{1}} & \textbf{COCO-20\textsuperscript{2}} & \textbf{COCO-20\textsuperscript{3}} & \textbf{Mean} \\
    \hline
    VGG 16 & B &  12.13 & 11.65 & 14.86  & 22.09 & 13.26 \\
    & B+C1 & 15.65  & 13.91 & 15.36  & 23.50 & 17.11 \\
    & B+C2 & 13.28 & 14.91 & 19.50  & 23.22 & 17.73 \\
    & B+C1+C2 & \textbf{18.35} & \textbf{16.72} & \textbf{19.59}  & \textbf{25.43} & \textbf{20.02} \\
    \hline
    ResNet 101 & B & 13.11  & 14.80 & 14.54 & 26.48  &  17.23 \\
    & B+C1 & 15.48  & 14.76 & 16.85  & 28.35 & 18.86 \\
    & B+C2 & 14.09  & 17.82 & 18.46  & 27.85 & 19.56 \\
    & B+C1+C2 & \textbf{16.98} & \textbf{17.98} & \textbf{20.96}  & \textbf{28.85} & \textbf{21.19} \\
    \hline
    \hline
  \end{tabular}
  \caption{The mIoU of ablations in the one-shot setting on COCO-20\textsuperscript{i}.}\label{tab:result_coco_one}
\end{table*}

\begin{table*}[t]
  \centering
  \begin{tabular}{l|l|cccc|c}
    \hline
    \hline
     \textbf{Backbone} & \textbf{Methods} & \textbf{COCO-20\textsuperscript{0}} & \textbf{COCO-20\textsuperscript{1}} & \textbf{COCO-20\textsuperscript{2}} & \textbf{COCO-20\textsuperscript{3}} & \textbf{Mean} \\
    \hline
    VGG 16 & Average & 20.76 & 16.87 & 20.55  & 27.61 & 21.45 \\
    ResNet 101 & Average & 18.73  & 18.46 & 21.27 & 29.20  &  21.92 \\
    \hline
    VGG 16 & Our-K-shot & \textbf{20.94}  & \textbf{19.24} & \textbf{21.94}  & \textbf{28.39} & \textbf{22.63} \\
    ResNet 101 & Our-K-shot & \textbf{19.13}  & \textbf{21.46} & \textbf{23.93}  & \textbf{30.08} & \textbf{23.65} \\
    \hline
    \hline
  \end{tabular}
  \caption{The mIoU of B+C1+C2 in the five-shot setting on COCO-20\textsuperscript{i}.}\label{tab:result_coco_five}
\end{table*}

{\bf Metrics.} As in  \cite{shaban2017one,rakelly2018conditional,dong2018few}, we use the mean intersection-over-union (mIoU) for quantitative evaluation. IoU of class $l$ is defined  as $\text{IoU}_l=\frac{TP_l}{TP_l + FP_l + FN_l}$, where $TP, FP$ and $FN$ are the number of pixels that are true positives, false positives and false negatives of the predicted segmentation masks, respectively. The mIoU is an average of the IoUs of different classes,  $\text{mIoU} = \frac{1}{n_l}\sum_l \text{IoU}_l$, where $n_l$ is the number of test classes. We report the mIoU averaged over the four folds of cross-validation.

{\bf Baselines, Ablations, and Variants of Our Approach.}
As a baseline B, we consider the approach depicted in the gray box ``PRIOR WORK'' in Fig.~\ref{fig:architecture} and specified in Sec.~\ref{sec:training} before our contribution 1. We also consider several ablations of our approach: B+C1 -- extends the baseline B with contribution 1 only; B+C2 -- extends the baseline B with contribution 2 only; and B+C1+C2 -- represents our full approach. These ablations are aimed at testing the effect of each of our contributions on performance. In addition, we consider two alternative neural networks -- VGG 16 and ResNet 101 -- as the CNN for extracting image features. VGG 16 has also been used in prior work \cite{shaban2017one,rakelly2018conditional,dong2018few}. We also compare with an approach called Upper-bound that represents a variant of our full approach B+C1+C2 trained such that both training and testing datasets consist of the same classes. As Upper-bound does not encounter new classes in testing, it represents an upper bound of our full approach. Finally, in the K-shot setting, we consider another baseline called Average. It represents our full approach B+C1+C2 that first independently predicts segmentations of the query image for each of the $K>1$ support images, and then averages all of the predictions. Our approach for the K-shot setting is called Our-K-shot, and differs from Average in that we rather jointly analyze all of the $K$ support images than treat them independently, as explained in Sec.~\ref{sec:k-shot}.

{\bf Training/testing time.}
The training/testing time is reported in Tab.~\ref{tab:time}. We can see that the contribution 1 just adds very small computational overhead over the baseline but significantly outperforms the baseline. Additionally, although contribution 2 has substantially larger testing time (about 40\% with VGG backbone and 35\% with ResNet backbone compare to the baseline), but it yields more significant performance gain than contribution 1 does.

{\bf One-shot Segmentation.}
Tab.~\ref{tab:ablation} compares our B+C1+C2 with the state of the art, ablations, and aforementioned variants in the one-shot setting on PASCAL-$5^i$. B+C1+C2 gives the best performance for both VGG 16 and ResNet 101, where the latter configuration significantly outperforms the state of the art with the increase in the mIoU averaged over the four folds of cross-validation by 13.49\%. Relative to B, our first contribution evaluated with B+C1 gives relatively modest performance improvements. From the results for B+C2, our second contribution produces larger gains in performance relative to B and B+C1, suggesting that contribution 2 in and of itself is more critical than contribution 1. Interestingly, combining both contribution 1 and contribution 2 significantly improves the results relative to using either contribution only. We also observe that performance of our B+C1+C2 for some folds of cross-validation (e.g., PASCAL-$5^2$ and PASCAL-$5^3$)  comes very close to that of Upper-bound, suggesting that our approach is very effective in generalizing to new classes in testing.  Fig.~\ref{fig:num_experts} shows the mIoU of B+C1+C2 as a function of the number of experts $N$ in the one-shot setting on PASCAL-$5^i$. As can be seen, for $N\ge 10$ our approach is not sensitive to a particular choice of $N$. We use $N=10$ as a good trade-off between complexity and accuracy.

\begin{figure}[h!]
    \centering
    \includegraphics[scale=0.3]{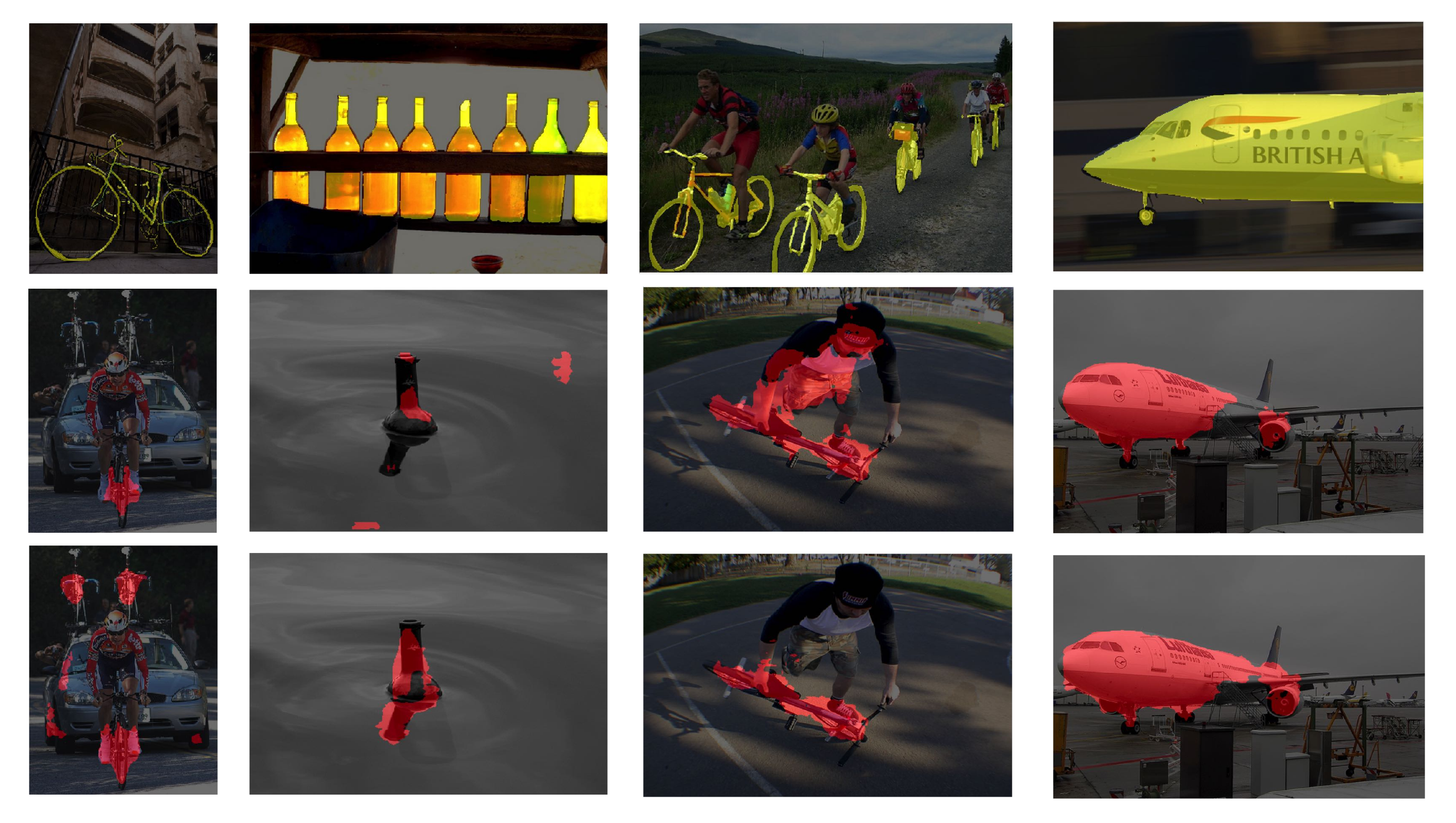}
    \caption{Examples from PASCAL-$5^0$. Top row: the support images with ground-truth segmentations in yellow. Middle row: the query images and segmentations in red predicted  by B+C1+C2 in the one-shot setting. Bottom row: the query images and segmentations in red predicted  by Our-K-shot in the five-shot setting (the remaining four support images are not shown). Our-K-shot in general improves performance over B+C1+C2, as  Our-K-shot effectively uses the higher level of supervision in  the five-shot setting. Best viewed in color.}
    \label{fig:qualitative2}
\end{figure}  

{\bf Five-shot Segmentation.}
Tab.~\ref{tab:five-shot} compares Our-K-shot with the state of the art and Average in the five-shot setting on PASCAL-$5^i$. Our-K-shot gives the best performance for both VGG 16 and ResNet 101, where the latter configuration significantly outperforms the state of the art with the increase in the mIoU averaged over the four folds of cross-validation by 15.97\%. In comparison with Average, the joint analysis of $K$ support images by Our-K-shot appears to be more effective, as Our-K-shot gives superior performance in every fold of cross-validation.

{\bf Results on COCO-$20^i$.}
 Tab.~\ref{tab:result_coco_one} and Tab.~\ref{tab:result_coco_five} shows our ablations' results in the one-shot and five-shot settings on COCO-$20^i$. The former results are obtained with B+C1+C2 and the latter, with Our-K-shot.
 The lower values of mIoU relative to those in Tab.~\ref{tab:ablation} and Tab.~\ref{tab:five-shot} indicate that COCO-$20^i$ is more challenging than PASCAL-$5^i$. Surprisingly, in fold COCO-$20^0$, B+C1+C2 with VGG 16 outperforms its counterpart with ResNet 101 in the one-shot setting. The same holds for Our-K-shot in the five-shot setting. On average, using ResNet 101 gives higher results. As expected, the increased supervision in the five-shot setting in general gives higher accuracy than the one-shot setting.

{\bf Qualitative Results.}
Fig.~\ref{fig:qualitative2} shows challenging examples from  PASCAL-$5^0$, and our segmentation results obtained with B+C1+C2 with ResNet 101 for the one-shot setting, and Our-K-shot with ResNet 101 for the five-shot setting. In the leftmost column, the bike in the support image has different pose from the bike in the query image. While this example is challenging for B+C1+C2, our performance improves when using Our-K-shot. In the second column from left, the query image shows a partially occluded target -- a part of the bottle. With five support images, Our-K-shot improves performance by capturing the bottle's shadow. The third column from left shows that the bike's features in the support image are insufficiently discriminative as the person also gets segmented along with the bike. With more examples, the bike is successfully segmented by Our-K-shot. In the rightmost column, the plane in the support image is partially occluded, and thus in the query image B+C1+C2 can only predict the head of the airplane while Our-K-shot's predicted segment covers most of the airplane.

\section{Conclusion}\label{sec:conclusion}

We have addressed one-shot and few-shot object segmentation, where the goal is to segment a query image, given a support image and the support's ground-truth segmentation. We have made two contributions. First, we have formulated an optimization problem that encourages high feature responses on the foreground and low feature activations on the background for more accurate object segmentation. Second, we have specified the gradient boosting of our model for fine-tuning to new classes in testing. Both contributions have been extended to the few-shot setting for segmenting the query by jointly analyzing the provided support images and their ground truths, rather than treating the support images independently. For evaluation, we have compared with prior work, strong baselines, ablations and variants of our approach on the PASCAL-$5^i$ and COCO-$20^i$ datasets. We significantly outperform the state of the art on both datasets and in both one-shot and five-shot settings. Using only the second contribution gives better results than using only the first contribution. Our integration of both contributions gives a significant gain in performance over each.

\noindent{\bf{Acknowledgement}}. This work was supported in part by DARPA XAI Award N66001-17-2-4029 and AFRL STTR AF18B-T002.
\pagebreak

\pagebreak

{\small
\bibliographystyle{ieee_fullname}
\bibliography{egbib}
}

\end{document}